\documentclass[a4paper,11pt]{article}

\usepackage{subcaption} 
\usepackage[english]{babel}
\usepackage[utf8]{inputenc}
\usepackage[T1]{fontenc}
\usepackage{amsmath, amssymb}
\usepackage{graphicx}
\usepackage{booktabs}
\usepackage{siunitx}
\usepackage{hyperref}
\usepackage{xcolor}
\usepackage{geometry}
\usepackage{amsmath}
\usepackage{hyperref}   

\title{
\rule{\linewidth}{1.2pt} \\[0.5em]
\textbf{PUB: A Plasma-Propelled Ultra-Quiet Blimp with Two-DOF Vector Thrusting}
\\[0.2em]
\rule{\linewidth}{1.2pt}
}

\author{Zihan Wang}
\date{August 2024}

\begin{document}
\maketitle

\begin{abstract}
This study presents the design and control of a Plasma-propelled Ultra-silence Blimp (PUB), a novel aerial robot employing plasma vector propulsion for ultra-quiet flight without mechanical propellers. The system utilizes a helium-lift platform for extended endurance and a four-layer ring asymmetric capacitor to generate ionic wind thrust. The modular propulsion units allow flexible configuration to meet mission-specific requirements, while a two-degree-of-freedom (DOF) head enables thrust vector control. A closed-loop slip control scheme is implemented for stable maneuvering. Flight experiments demonstrate full-envelope capability, including take-off, climb, hover, descent, and smooth landing, confirming the feasibility of plasma vector propulsion, the effectiveness of DOF vector control, and the stability of the control system. Owing to its low acoustic signature, structural simplicity, and high maneuverability, PUB is well suited for noise-sensitive, enclosed, and near-space applications.
\end{abstract}

\section{Introduction}
In 2024, the ``low-altitude economy'' was written into China's Government Work Report for the first time~\cite{govreport2024}, and flying robots have been rapidly popularized nationwide. From an environmental perspective, electrically powered air vehicles are attracting growing attention; key technologies include overall configuration design, integrated energy management, and high-efficiency, high power-to-weight electric propulsion~\cite{whitepaper2019}. For electric propulsion, mainstream systems use electric motors to drive propellers, but propeller noise is significant and hard to mitigate~\cite{low_noise_propeller}, which limits widespread use in cities---the main arena for the low-altitude economy---and is also unfavorable for silent reconnaissance. Hence, there is a pressing need for a new propulsion approach enabling quiet, fully electric flight.

In the 1920s, Brown observed that an asymmetric capacitor under high voltage can generate thrust, known as the Biefeld--Brown effect. A leading explanation is \emph{ionic wind}: a high electric field ionizes air, and the resulting ions accelerate and transfer momentum to neutral molecules, producing a net airflow (thrust)~\cite{brown_effect_analysis}. Xu \emph{et al.} first mounted a plasma thruster on a fixed-wing UAV without other propulsion; the gliding distance with the thruster on was five times that with it off, but the maximum range was only \SI{45}{m} and no controller design was provided~\cite{xu2018nature}. Zhang \emph{et al.} realized altitude control for a micro ionic-wind-powered UAV using passive components, but the wingspan was at most \SI{6.3}{cm}, limiting applicability~\cite{zhang2023cjoa}.

We propose PUB, a plasma-propelled ultra-quiet blimp. An ellipsoidal helium balloon provides buoyancy, and a two-DOF gimbal is suspended beneath the envelope. A ring-type plasma thruster is mounted on the gimbal; by controlling gimbal angles, the thrust direction is vectored. The thruster uses a four-layer concentric-ring asymmetric electrode structure. A high-voltage supply evokes the Biefeld--Brown effect; thrust is modulated by gimbal orientation and high-voltage level.

\section{Conceptual Design}
\subsection{Overall Layout}
\subsubsection{Blimp Design}
\begin{figure}[htbp]
  \centering
  \begin{subfigure}{0.4\textwidth}
    \centering
    \includegraphics[width=\linewidth]{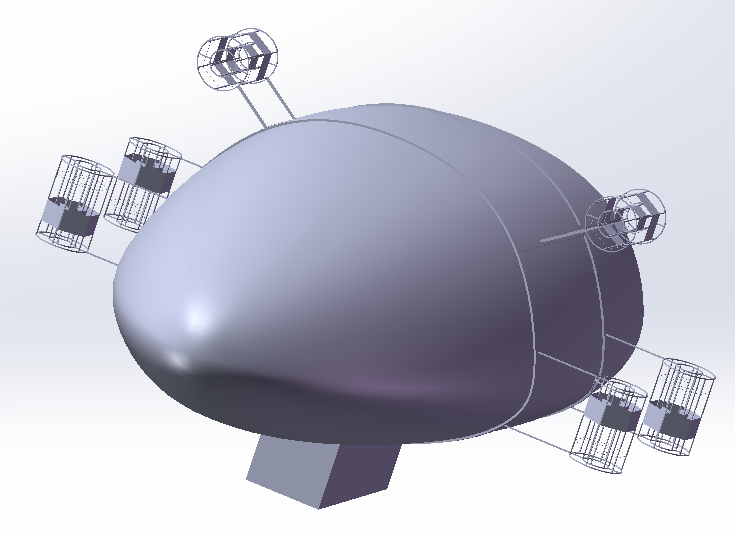}
    \caption{PUB, first-generation overall layout.}
    \label{fig:fig1}
  \end{subfigure}
  \hspace{0.05\textwidth} 
  \begin{subfigure}{0.47\textwidth}
    \centering
    \includegraphics[width=\linewidth]{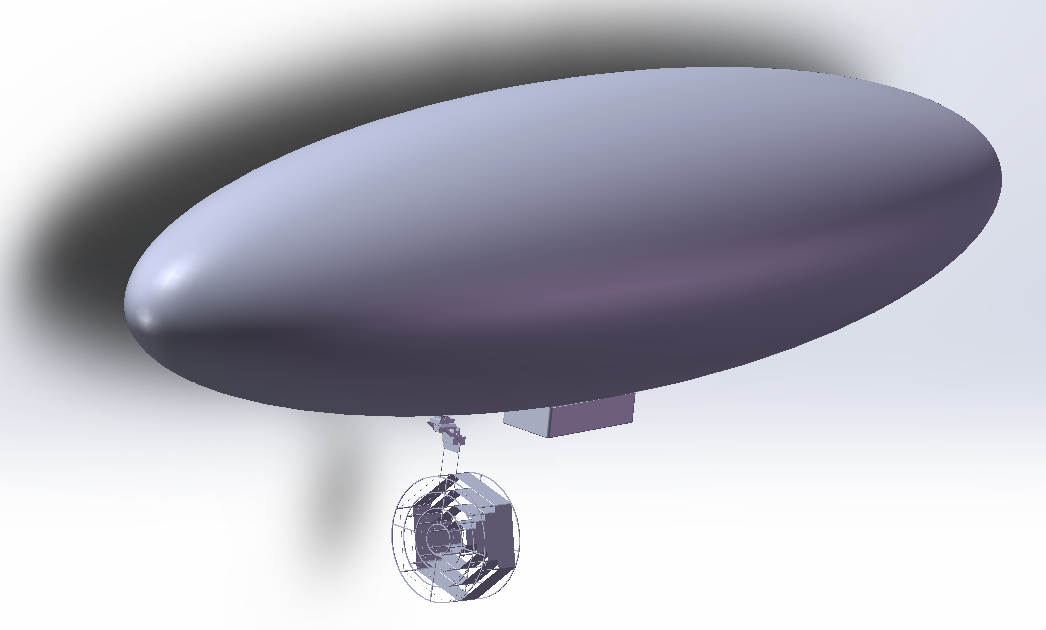}
    \caption{PUB, second-generation overall layout.}
    \label{fig:fig2}
  \end{subfigure}
\end{figure}

As shown in Fig.~\ref{fig:fig1}, the main body of PUB is an ellipsoidal aluminum-film helium balloon of length \SI{1.97}{m} and maximum diameter \SI{51}{cm}. The volume is approximated by the ellipsoid formula
\begin{equation}
  V = \frac{4}{3}\pi a b c,
\end{equation}
which yields $V \approx \SI{268.29}{L}$. Each liter of helium provides approximately \SI{1.11}{g} of buoyant lift, giving a maximum gross lift of about \SI{297.8}{g}. The envelope mass is measured as \SI{79.36}{g}, so the net available lift is approximately \SI{218.44}{g}.

A lightweight 3D-printed gondola is rigidly attached under the aerostat to host the electrical and flight-control modules and to concentrate the payload.

The thruster layout has undergone several iterations. In the first-generation design (Fig.~\ref{fig:fig1}), six identical thrusters are rigidly attached at different positions, requiring simultaneous control of their on/off states and thrust magnitudes to realize six-DOF motion. This resulted in high system mass and control complexity. The second-generation design (Fig.~\ref{fig:fig2}) simplifies the layout to a \emph{single} thruster mounted on a two-DOF platform directly beneath the envelope, enabling thrust vectoring in pitch and yaw and reducing weight while enhancing controllability.

\subsubsection{Two-DOF Gimbal}
\begin{figure}[htbp]
  \centering
  \includegraphics[width=0.45\textwidth]{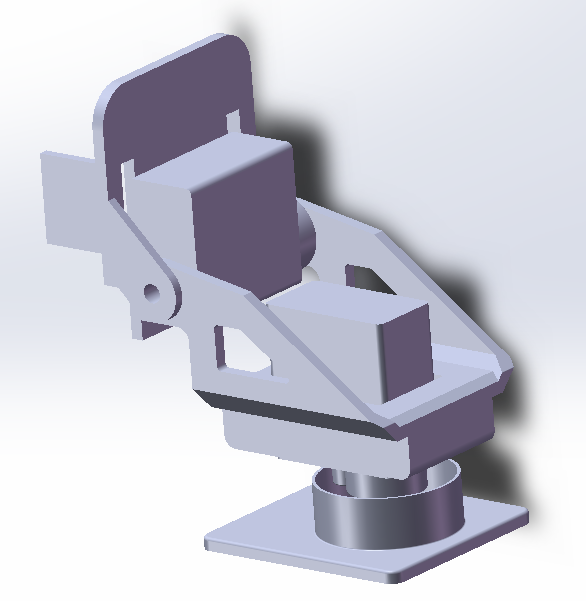}
  \caption{Two-DOF gimbal.}
  \label{fig:fig3}
\end{figure}

As shown in Fig.~\ref{fig:fig3}, the gimbal uses two RC servos (each \SI{9}{g}; total gimbal mass $\approx$\SI{33}{g}; rated torque \SI{1.6}{kg.cm}) to vector the thruster. The upper stage rotates in the blimp's $oxz$ symmetry plane to control pitch, with an angular range of $0^\circ$--$180^\circ$. The lower stage is rigidly attached beneath the blimp and rotates about the $z$ axis to control yaw, also within $0^\circ$--$180^\circ$.

\subsection{Plasma Thruster}
\begin{figure}[htbp]
  \centering
  \includegraphics[width=0.45\textwidth]{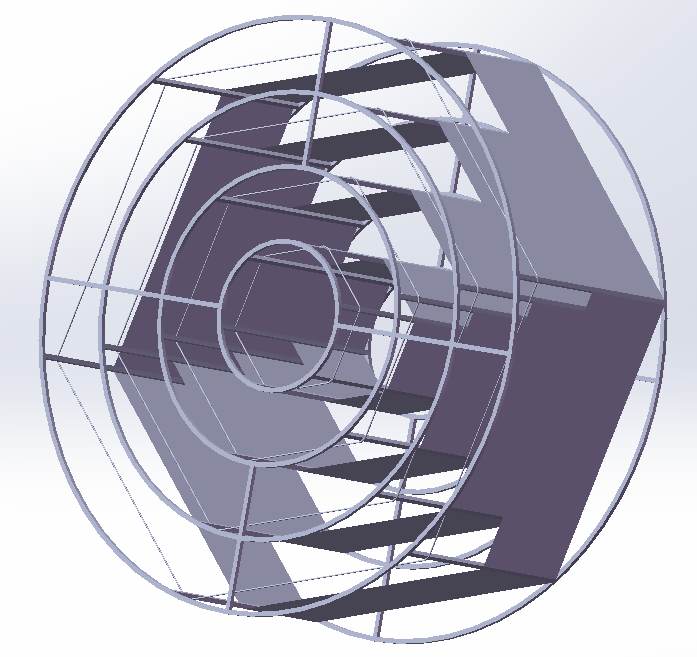}
  \caption{Four-layer concentric-ring plasma thruster.}
  \label{fig:fig4}
\end{figure}

Fig.~\ref{fig:fig4} shows the thruster: a four-layer concentric ring frame with \SI{9.44}{mm} spacing between adjacent rings. The positive electrode is \SI{0.1}{mm} copper wire; the negative electrode is \SI{40}{mm}-wide aluminum foil. The electrode separation is \SI{25}{mm}.

\subsection{Electrical System}
\begin{figure}[htbp]
  \centering
  \includegraphics[width=0.7\textwidth]{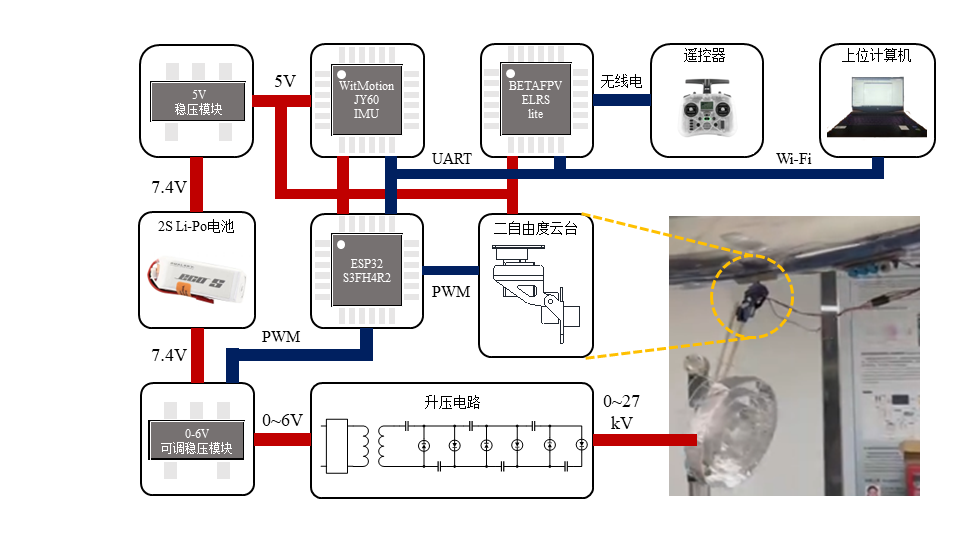}
  \caption{Electrical modules of PUB.}
  \label{fig:fig5}
\end{figure}

The electrical system (Fig.~\ref{fig:fig5}) is weight-optimized due to limited buoyant lift and includes only essential modules for propulsion and control. An ESP32-S3 board serves as the main controller, communicating via Wi-Fi to a ground station. A BETAFPV ELRS Lite receiver enables RC control within \SI{500}{m}. An IMU module (e.g., JY60) provides attitude sensing with embedded filtering. Power is supplied by a 2S \SI{800}{mAh} Li-ion battery (\SI{42.5}{g}). A regulator feeds the controller and the HV stage, with the latter modulated by PWM in the \SIrange{0}{6}{V} range. The high-voltage topology follows a DC source $\rightarrow$ pre-boost inverter $\rightarrow$ transformer $\rightarrow$ multi-stage voltage doubler, yielding about \SI{27}{kV} for the thruster~\cite{xu_thesis_2020}.

\paragraph{Main Parameters.}
Table~\ref{tab:params} lists the key parameters measured on the prototype.

\begin{table}[htbp]
  \centering
  \caption{Main parameters of PUB.}
  \label{tab:params}
  \begin{tabular}{@{}ll@{}}
    \toprule
    \textbf{Parameter} & \textbf{Value} \\ \midrule
    \multicolumn{2}{@{}l}{\textit{Aluminum-Film Balloon}} \\
    Mass & \SI{79.36}{g} \\
    Length & \SI{1.97}{m} \\
    Max diameter & \SI{51}{cm} \\ \addlinespace
    \multicolumn{2}{@{}l}{\textit{Two-DOF Gimbal}} \\
    Mass (each servo) & \SI{9}{g} \\
    Total mass & \SI{33}{g} \\
    Servo torque & \SI{1.6}{kg.cm} \\ \addlinespace
    \multicolumn{2}{@{}l}{\textit{Thruster}} \\
    Ring spacing (adjacent) & \SI{9.44}{mm} \\
    Positive electrode diameter & \SI{0.1}{mm} \\
    Negative electrode width & \SI{40}{mm} \\ \addlinespace
    \multicolumn{2}{@{}l}{\textit{Electrical}} \\
    Battery capacity & \SI{800}{mAh} \\
    Battery mass & \SI{42.5}{g} \\
    HV output (approx.) & \SI{27}{kV} \\
    \bottomrule
  \end{tabular}
\end{table}

\section{Theory and Experiments of the Plasma Thruster}
\subsection{Theory of the Plasma Thruster}
\subsubsection{Plasma--Neutral Interaction Model}
In a high-voltage electric field, corona discharge occurs in the air. Gas molecules are ionized by the strong electric field, producing ions. These ions are subject to the external electric field between the plates and interact with neutral gas molecules through collisions in the air flow, exchanging momentum and energy. This coupling effect accelerates the air around the corona discharge electrode, generating thrust~\cite{taccogna2004}. A simplified general model will be constructed to describe the interaction between the ionized plasma and neutral air molecules and its role in the flow.

Considering the actual operating conditions of the ionic-wind thruster, we can make the following assumptions to simplify the analysis: assume that the number of ions in the flow is sufficiently sparse, so ions can be considered to interact only with neutral gas molecules through collisions. There is no momentum exchange between ions due to thermal motion, no stress, and only diffusion. The electric field strength in the studied interaction region is not excessively large, and the accelerating effect of the external electric field on the ions has little effect on the deviation of the ion velocity spatial probability distribution, so a first-order approximation can be used. The intensity of the electric field's effect on the ions is much greater than the inertia of the ion motion, so the change in momentum carried by the ions themselves is ignored.

First, we analyze the momentum exchange relationship between the charged plasma and neutral air molecules. When the ion flow has a slip velocity relative to the local neutral particle flow, the force exerted by the ion flow micro-parcel on the neutral molecule micro-parcel can be given by molecular kinetic theory, approximately using the steel ball collision model~\cite{hinton1983}:

\begin{equation}
\begin{aligned}
f &= \iiint \iiint \Omega_{D} |\boldsymbol{v}_{air} - \boldsymbol{v}_{ion}|
      \frac{4}{3} \frac{mM}{m+M} (\boldsymbol{v}_{air} - \boldsymbol{v}_{ion}) \\
  &\quad \cdot n_{air} \left( \frac{M}{2\pi kT} \right)^{\frac{3}{2}}
      e^{-\frac{M |\boldsymbol{v}_{air} - \boldsymbol{u}|^{2}}{2kT}}
      d\boldsymbol{v}_{airx} \, d\boldsymbol{v}_{airy} \, d\boldsymbol{v}_{airz} \\
  &\quad \cdot n_{ion} \left( \frac{m}{2\pi kT} \right)^{\frac{3}{2}}
      e^{-\frac{m |\boldsymbol{v}_{ion}|^{2}}{2kT}}
      d\boldsymbol{v}_{ionx} \, d\boldsymbol{v}_{iony} \, d\boldsymbol{v}_{ionz}
\end{aligned}
\end{equation}

The integration result is:

\begin{equation}
f = \frac{64}{9} \Omega_{D} \sqrt{\frac{kT}{2\pi}} 
    \sqrt{\frac{mM}{m+M}} n_{ion} n_{air} \boldsymbol{u}
\end{equation}

Ignoring the inertia of the ions themselves, we approximate that the ions maintain equilibrium under the momentum exchange from collisions with neutral particles and the electric field force:

\begin{equation}
\boldsymbol{E} (n_{ion} q_{ion}) = \boldsymbol{f}
= \frac{64}{9} \Omega_{D} \sqrt{\frac{kT}{2\pi}}
  \sqrt{\frac{mM}{m+M}} n_{ion} n_{air} \boldsymbol{u}
\end{equation}

Define the mobility:

\begin{equation}
\mu_{i} = \frac{9}{64} \frac{q_{ion}}{n_{air}}
          \sqrt{\frac{2\pi}{kT}}
          \sqrt{\frac{1}{m} + \frac{1}{M}} \, \Omega_{D}
\end{equation}

According to the Einstein relation between mobility and diffusivity:

\begin{equation}
D_{i} = \mu_{i} \left( \frac{kT}{q_{ion}} \right)
\end{equation}

The current flow caused by the diffusion coefficient is:

\begin{equation}
\boldsymbol{j}_{d} = -D_{i} \nabla (n_{ion} q_{ion})
\end{equation}

The ionization and recombination of plasma back to neutral state are simplified as source terms $S_{ion} - L_{ion}$. Using $\rho_{e}$ to re-characterize the spatial distribution of charged particles, the governing equations for the flow can be summarized by combining Maxwell's equations and the N-S equations for incompressible fluids:

\begin{equation}
\left\{
\begin{aligned}
&\nabla \cdot \boldsymbol{E} = \rho_{e} / \varepsilon, \\
&\nabla \times \boldsymbol{E} = 0, \\
&\frac{\partial \rho_{e}}{\partial t} + \nabla \cdot \boldsymbol{j} = S_{ion} - L_{ion}, \\
&\boldsymbol{j} = \rho_{e} (\boldsymbol{v} + \mu_{i} \boldsymbol{E}) - D_{i} \nabla \rho_{e}, \\
&\nabla \cdot \boldsymbol{v} = 0, \\
&\frac{\partial \boldsymbol{v}}{\partial t} + (\boldsymbol{v} \cdot \nabla) \boldsymbol{v} 
  = -\frac{1}{\rho} \nabla p + \nu \nabla^{2} \boldsymbol{v} + \frac{\rho_{e}}{\rho} \boldsymbol{E}
\end{aligned}
\right.
\end{equation}

\subsubsection{Principle of Ionic-Wind Propulsion}
\begin{figure}[htbp]
  \centering
  \includegraphics[width=0.55\textwidth]{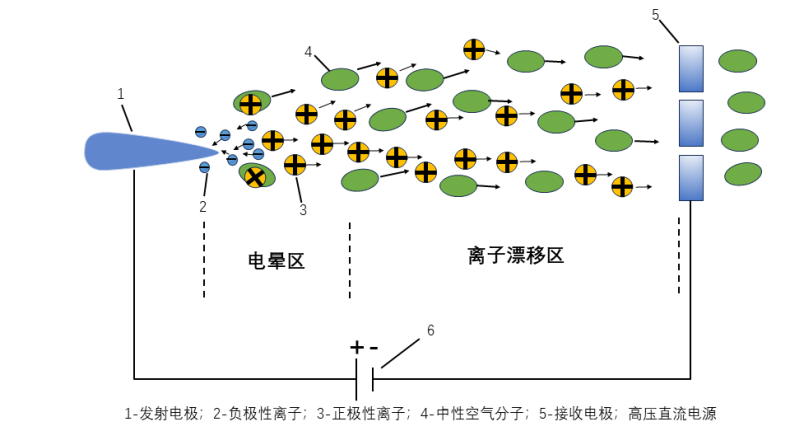}
  \caption{Schematic of corona discharge for ionic-wind generation.}
  \label{fig:fig6}
\end{figure}

\paragraph{Biefeld--Brown Effect.}
The effect states that a properly shaped electrode pair immersed in a dielectric and biased at high voltage generates a net force (Fig.~\ref{fig:fig6}). 

The ionic wind generation device mainly consists of two major components: the emitter electrode and the collector electrode. When a voltage is applied between the two electrodes, a non-uniform electric field is formed. The tip of the emitter electrode concentrates space charges, and the electric field strength near the tip becomes particularly high, making it more prone to corona discharge ~\cite{wilson2009}. As the voltage increases, it reaches a critical value sufficient to ionize the gas near the emitter electrode. The gas near the tip first undergoes a local self-sustaining discharge, which produces a corona. The voltage at which the corona first appears is called the corona onset voltage. The air near the tip of the positively charged emitter electrode is ionized into positive ions and electrons. This area, known as the corona region, appears as a thin bluish-purple layer that expands as the voltage increases, caused by excited atoms returning to their ground state during the ionization process. Under the influence of the electric field, the electron avalanche in the corona region transitions into a streamer. Most of the ions with the same polarity as the emitter electrode accelerate toward the collector electrode. This region of ion acceleration and migration is called the ion drift region, where the electric field is relatively weak, and little ionization occurs. Current conduction in this region relies on the migration of positive ions. As the charged ions accelerate, they collide with neutral air molecules, transferring momentum and creating a macroscopic, directional airflow. This airflow is known as ionic wind, or corona wind ~\cite{chang1991}.

\begin{figure}[htbp]
  \centering
  \includegraphics[width=0.55\textwidth]{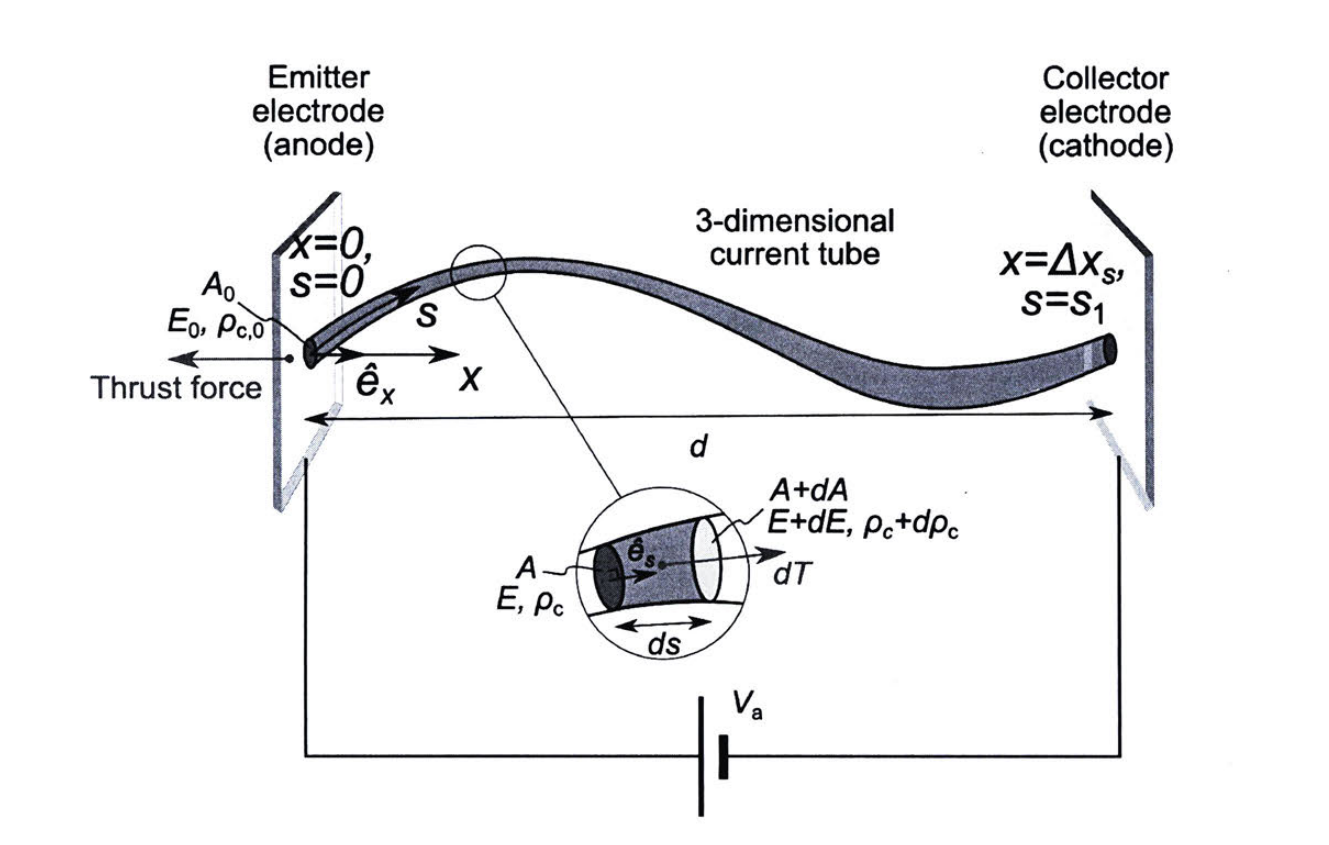}
  \caption{Schematic of thrust generation in a drift region.}
  \label{fig:fig7}
\end{figure}

\paragraph{Thrust Model.}
Based on the ion wind effect model from section 3.1.1, we can further simplify and form a thrust model for the ion propulsion device~\cite{ahedo2011}~\cite{xu_thesis_2020}. Fig.~\ref{fig:fig7} illustrates the thrust direction.

Ignoring the effect of air viscosity on the electrodes, the thrust of an ion wind thruster is simply the reaction force of the charges in the flow field on the electric field:
\begin{equation}
\mathbf{T} = \iiint -\rho_{\mathrm{e}} \mathbf{E} \, dV
\end{equation}
where $\rho_e$ is space charge density and $\mathbf{E}$ the electric field.

Introducing a new approximate assumption to simplify the model: assume $\mathbf{u} \gg \mathbf{v}$, neglect the macroscopic motion velocity of the neutral particle flow field, i.e., neglect the current caused by ions moving with the macroscopic neutral flow; neglect the non-uniform distribution of ions and the current caused by diffusion.
\begin{equation}
\mathbf{j} = \rho_{\mathrm{e}} \mathbf{u} = \rho_{\mathrm{e}} \mu_{\mathrm{i}} \mathbf{E}
\end{equation}
Under stable operating conditions of the thruster, the flow is steady $\frac{\partial \rho_{\mathrm{e}}}{\partial t} = 0$; in the main ion drift region generating thrust, there is no ionization or recombination, so the source term $S_{\mathrm{ion}} - L_{\mathrm{ion}} = 0$:

\begin{equation}
\nabla \cdot \mathbf{j} = 0
\end{equation}

Thus, we can analyze a certain flow tube of the plasma flow, denoting the cross-sectional area as $A$:

\begin{equation}
\begin{aligned}
d\mathbf{T} &= -\rho_{\mathrm{e}} \mathbf{E} \cdot A \, ds \\
            &= -\rho_{\mathrm{e}} \frac{\mathbf{u}}{\mu_{\mathrm{i}}} \cdot A \, ds \\
            &= -\frac{\mathbf{j} \, A}{\mu_{\mathrm{i}}} \, ds
\end{aligned}
\end{equation}

The flux along the flow tube remains constant $|\mathbf{j} A| = d I$, with direction along $ds$, and the total thrust can be obtained by integration:

\begin{equation}
\mathbf{T} = -\int \frac{d I}{\mu_{\mathrm{i}}} \int d\mathbf{s}
\end{equation}
The current flows from the positive electrode to the negative electrode, so $\int d\mathbf{s} = \mathbf{l}$, where $\mathbf{l}$ is the vector pointing from the positive electrode to the negative electrode:

\begin{equation}
\mathbf{T} = -\frac{\mathbf{l} \, I}{\mu_{\mathrm{i}}}
\end{equation}

\subsection{Experiments on the Plasma Thruster}
\subsubsection{Thrust Measurement}
The thruster is the core power source for the entire PUB flying robot. Based on key requirements such as a simplified, lightweight structure and a high thrust-to-weight ratio, we went through three design iterations for the thruster: single-ring, dual-ring, and quad-ring.

For the thruster's material, we needed to consider insulation, low weight, and high toughness. We chose Ledo 6060 SLA photosensitive resin as the 3D printing material. After attaching aluminum foil and wrapping copper wires, the quad-ring thruster weighed only \SI{19.64}{\gram}, which is very light and met our weight budget. An insulation test showed excellent insulating properties, creating ideal conditions for plasma transmission between the electrodes.

\begin{figure}[htbp]
  \centering
  \includegraphics[width=0.55\textwidth]{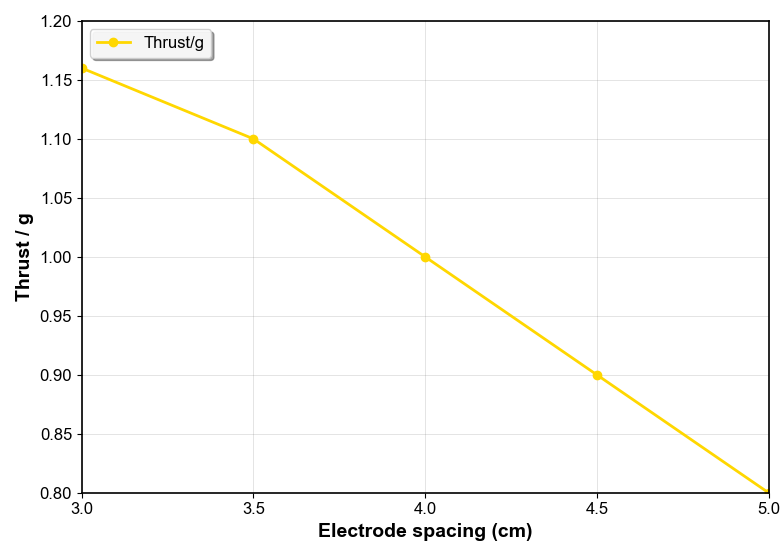}
  \caption{Thrust vs.\ Electrode spacing.}
  \label{fig:fig8}
\end{figure}

According to the plasma propulsion principles in section 3.1.2, the distance between the positive copper wire and the negative aluminum foil affects the thrust. To find the optimal spacing, we repeatedly changed the gap of the dual-ring thruster and measured the thrust. The experimental process was as follows: The thruster was inverted on a high-precision electronic scale (\SI{0.01}{\gram} accuracy) with the negative electrode facing up, ensuring it was well-insulated from the scale. After taring the scale, the reading displayed the maximum thrust in grams when the power was turned on. The dual-ring thruster had a net weight of \SI{16.00}{\gram}. We conducted five parallel experiments, decreasing the electrode spacing each time. The thrust measurements are shown in Fig.~\ref{fig:fig8}. We found that as the spacing decreased, the thrust gradually increased, reaching a maximum thrust-to-weight ratio of \SI{0.7105}{\newton\per\kilogram}. However, when the gap was reduced to \SI{2.5}{\centi\meter}, a tip discharge phenomenon occurred, causing the negative aluminum foil to be punctured and burned. This demonstrated that designing the electrode spacing is a critical step.

Based on the thrust measurement experiments and the data in  Fig.~\ref{fig:fig8}, we concluded the following: when the spacing is greater than \SI{2.5}{\centi\meter}, a smaller gap increases the charge density and electric field strength, thereby increasing thrust. However, when the spacing is reduced to around \SI{2.5}{\centi\meter}, the tip discharge from the positive electrode becomes too intense, creating an overly strong electric field that punctures the aluminum foil.

For the final design, we improved the thruster to the four-ring structure shown in Fig.~\ref{fig:fig4} to achieve greater thrust. This version has a net weight of \SI{19.64}{\gram}. With an electrode spacing of \SI{3.0}{\centi\meter}, the measured maximum thrust was \SI{0.051}{\newton}. This gives the thruster a maximum thrust-to-weight ratio of \SI{2.597}{\newton\per\kilogram}, a significant improvement over the previous dual-ring version.

\subsubsection{Continuously Adjustable Thrust}
\begin{figure}[htbp]
  \centering
  \includegraphics[width=0.55\textwidth]{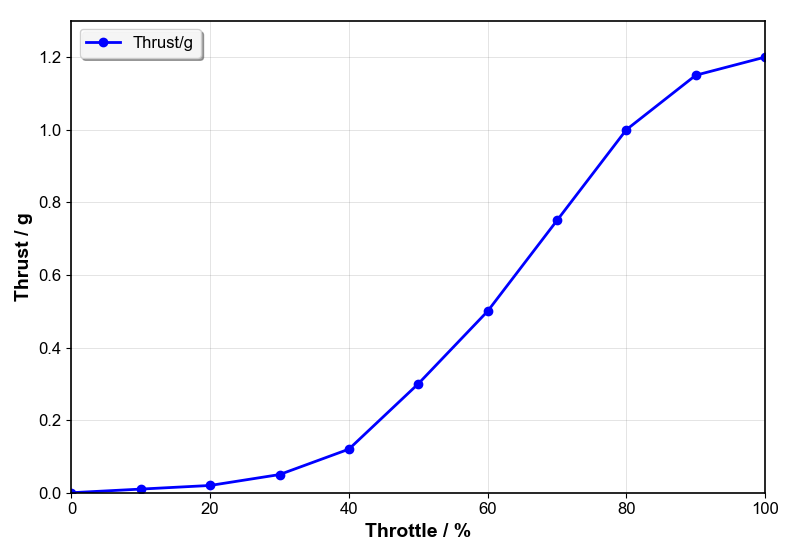}
  \caption{Thrust vs.\ Throttle setting.}
  \label{fig:fig9}
\end{figure}

\begin{figure}[htbp]
  \centering
  \includegraphics[width=0.55\textwidth]{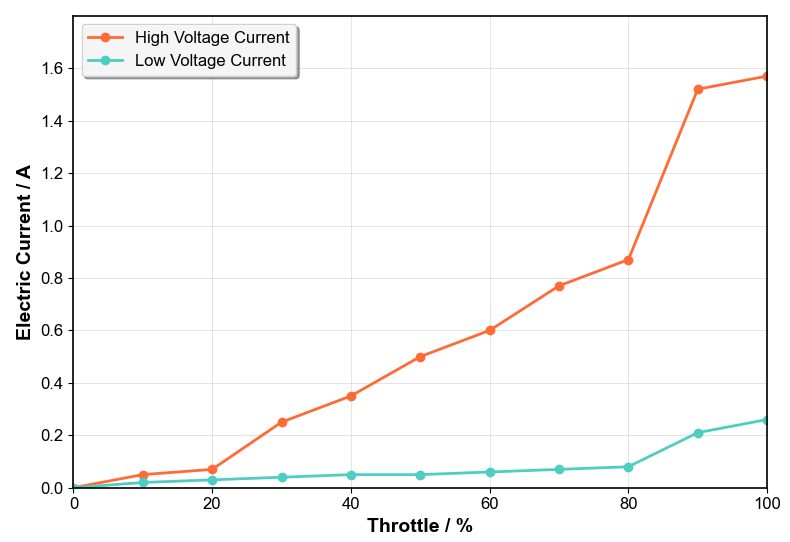}
  \caption{HV/LV current vs.\ T       hrottle setting.}
  \label{fig:fig10}
\end{figure}

By commanding throttle via RC, the high-voltage drive is modulated and the thruster produces approximately linear thrust vs.\ throttle, while the HV-side current changes more than the LV-side current. Fig.~\ref{fig:fig9} and~\ref{fig:fig10} show representative relationships.

Table~\ref{tab:influences} summarizes single-variable tests on factors affecting thrust.

\begin{table}[htbp]
  \centering
  \caption{Factors affecting thrust magnitude (illustrative data).}
  \label{tab:influences}
  \begin{tabular}{@{}lll@{}}
    \toprule
    \textbf{Variable} & \textbf{Value} & \textbf{Measured thrust (g)} \\ \midrule
    Electrode spacing & \SI{5.0}{cm} & 0.80 \\
    & \SI{4.5}{cm} & 0.90 \\
    & \SI{4.0}{cm} & 0.99 \\
    & \SI{3.5}{cm} & 1.10 \\
    & \SI{3.0}{cm} & 1.16 \\
    & \SI{2.5}{cm} & (foil puncture) \\ \addlinespace
    Throttle & 100\% & 1.20 \\
    & 90\% & 1.16 \\
    & 80\% & 0.70 \\
    & 70\% & 0.58 \\
    & 60\% & 0.45 \\
    & 50\% & 0.31 \\
    & 40\% & 0.12 \\
    & 30\% & 0.06 \\
    & 20\% & 0.00 \\ \addlinespace
    Copper wire diameter & \SI{0.2}{mm} & 0.73 \\
    & \SI{0.1}{mm} & 0.80 \\
    \bottomrule
  \end{tabular}
\end{table}

\section{Dynamic Modeling of the System}
With the gondola and thruster placed near the symmetry plane, PUB can be idealized as a rigid airship with one symmetry plane. At low speeds and with relatively small thruster deflections, rigid-body assumptions hold. Compared to conventional aircraft, thrust-vector direction and aerodynamics are special here.

\subsection{Reference Frames}
Define the inertial (ground) frame, body frame, and airflow frame. Within the scope of our study, wind speed is neglected, so the ground speed can be considered equal to the airspeed. 

The transformation between the ground frame and the body frame is defined by the attitude angles:

\begin{equation}
\boldsymbol{L}_{bg}=
\begin{bmatrix}
1 & 0 & 0\\
0 & \cos\phi & \sin\phi\\
0 & -\sin\phi & \cos\phi
\end{bmatrix}
\begin{bmatrix}
\cos\theta & 0 & -\sin\theta\\
0 & 1 & 0\\
\sin\theta & 0 & \cos\theta
\end{bmatrix}
\begin{bmatrix}
\cos\psi & \sin\psi & 0\\
-\sin\psi & \cos\psi & 0\\
0 & 0 & 1
\end{bmatrix}
\end{equation}

\begin{figure}[htbp]
  \centering
  \includegraphics[width=0.5\textwidth]{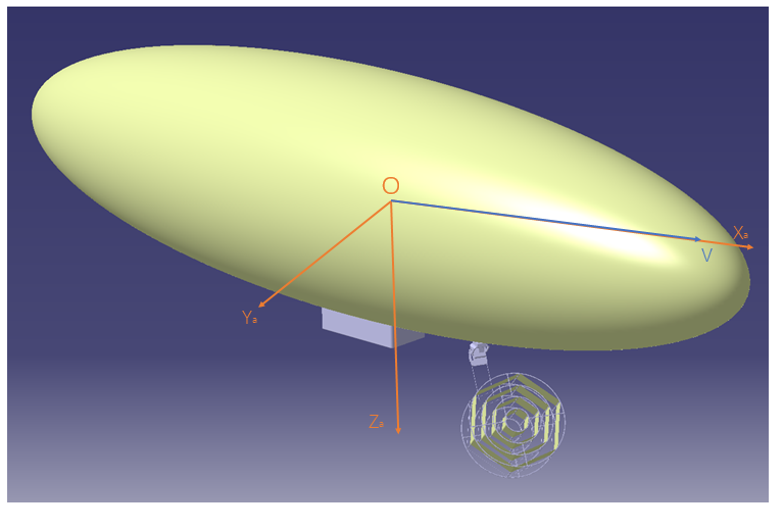}
  \caption{Definition of airflow-related coordinate frames.}
  \label{fig:fig11}
\end{figure}

Since the aerodynamics are simplified to the characteristics of an axisymmetric shape, a new airflow frame is defined: $x_a$ is along the freestream direction, $z_a$ lies in the $x_bx_a$ plane and is perpendicular to $x_a$, and $y_a$ is determined according to the right-hand rule. The angle of attack $\alpha$ is defined as the angle between the airship axis $x_b$ and the airflow frame $x_a$, and the sideslip angle $\beta$ is defined as the angle between the $x_bx_a$ plane of the freestream and the airship symmetry plane $x_bz_b$. Thus, the transformation from the airflow frame to the body frame is Fig.~\ref{fig:fig11}:

\begin{equation}
\boldsymbol{L}_{ba}=
\begin{bmatrix}
1 & 0 & 0\\
0 & \cos\beta & -\sin\beta\\
0 & \sin\beta & \cos\beta
\end{bmatrix}
\begin{bmatrix}
\cos\alpha & 0 & \sin\alpha\\
0 & 1 & 0\\
-\sin\alpha & 0 & \cos\alpha
\end{bmatrix}
\end{equation}

\subsection{External Forces on the Blimp}
\paragraph{Buoyancy.}Study the buoyancy and its moment effects in the body frame, denoting the distance from the center of buoyancy to the center of mass as d:

\begin{equation}
\mathbf{F}_f = \mathbf{L}_{bg}
\begin{bmatrix}
0 \\
0 \\
-G
\end{bmatrix},
\quad
\mathbf{M}_f =
\begin{bmatrix}
0 \\
0 \\
-d
\end{bmatrix}
\times
\mathbf{L}_{bg}
\begin{bmatrix}
0 \\
0 \\
-G
\end{bmatrix}
\end{equation}

\paragraph{Aerodynamics.} In the airflow frame, neglect the aerodynamic effects of the thruster and any payloads suspended below the airship, and consider the airship as an axisymmetric ellipsoid for aerodynamic effects. Lift is $L = L(\alpha)$, drag is $D = D(\alpha)$, and aerodynamic moment is $M = M(\alpha)$. Compared with the restoring moment generated by buoyancy, the damping moments from the airflow can be neglected.

\begin{equation}
\boldsymbol{F}_a = \begin{bmatrix}-D\\ \mathbf{0}\\ -L\end{bmatrix}, \quad
\boldsymbol{M}_a = \begin{bmatrix}0\\ M\\ 0\end{bmatrix} 
\end{equation}

\begin{figure}[htbp]
  \centering
  \includegraphics[width=0.7\textwidth]{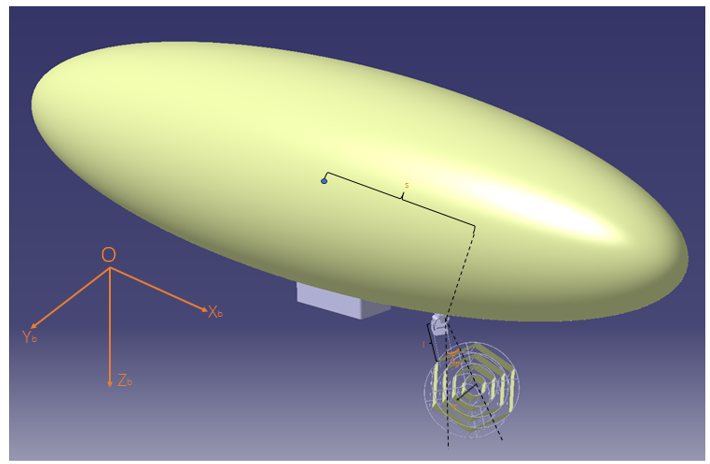}
  \caption{Thruster vectoring in pitch (top) and yaw (bottom).}
  \label{fig:fig12}
\end{figure}

\begin{figure}[htbp]
  \centering
  \includegraphics[width=0.75\textwidth]{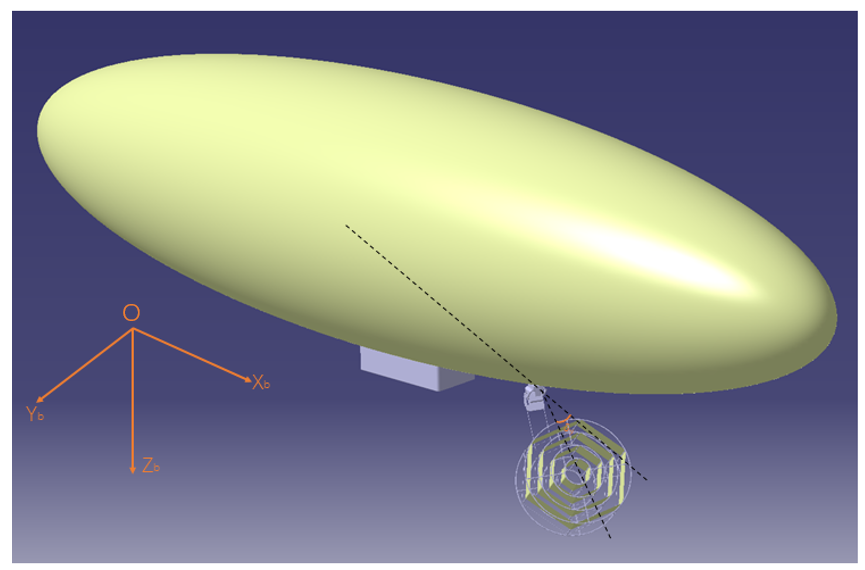}
  \caption{Transfer function model of controller.}
  \label{fig:fig13}
\end{figure}

\paragraph{Thruster.} Study the thrust effects of the thruster in the body frame(Fig.~\ref{fig:fig12},~\ref{fig:fig13}). Denote the yaw deflection angle introduced by the servo as $\delta_y$, the pitch deflection angle as $\delta_p$, the thruster installation position as $s$, and the thruster link length as $l$. Then the thrust force and moment generated by the thruster are:

\begin{equation}
\boldsymbol{F}_T =
\begin{bmatrix}
T \cos\delta_y \cos\delta_p\\
T \sin\delta_y \cos\delta_p\\
- T \sin\delta_p
\end{bmatrix}, \quad
\boldsymbol{M}_T =
\left(
\begin{bmatrix}s_x\\0\\s_z\end{bmatrix} +
\begin{bmatrix}
l \cos\delta_y \sin\delta_p\\
l \sin\delta_y \sin\delta_p\\
l \cos\delta_p
\end{bmatrix}
\right)
\times
\begin{bmatrix}
T \cos\delta_y \cos\delta_p\\
T \sin\delta_y \cos\delta_p\\
- T \sin\delta_p
\end{bmatrix} 
\end{equation}

\subsection{Equations of Motion}
\subsubsection{Dynamic Equations}
Choose the body frame to establish the translational dynamics about the center of mass: 

\begin{equation}
\begin{bmatrix}\dot{u}\\\dot{v}\\\dot{w}\end{bmatrix} =
\begin{bmatrix} vr - wq \\ -ur + wp \\ uq - vp \end{bmatrix} +
\frac{1}{m} \left(
\boldsymbol{L}_{ba} \begin{bmatrix}-D\\ \boldsymbol{0} \\ -L\end{bmatrix} +
\begin{bmatrix} T \cos\delta_y \cos\delta_p \\ T \sin\delta_y \cos\delta_p \\ -T \sin\delta_p \end{bmatrix}
\right)  
\end{equation}

Similarly, in the body frame, establish the rotational dynamics about the center of mass:

\begin{equation}
\begin{gathered}
\begin{bmatrix}0\\0\\-d\end{bmatrix}\times\boldsymbol{L}_{bg}\begin{bmatrix}0\\0\\-G\end{bmatrix}
+\boldsymbol{L}_{ba}\begin{bmatrix}0\\M\\0\end{bmatrix}
+\left(\begin{bmatrix}s_x\\0\\s_z\end{bmatrix}+\begin{bmatrix}l\cos\delta_y\sin\delta_p\\l\sin\delta_y\sin\delta_p\\l\cos\delta_p\end{bmatrix}\right)
\times\begin{bmatrix}T\cos\delta_y\cos\delta_p\\T\sin\delta_y\cos\delta_p\\-T\sin\delta_p\end{bmatrix}\\
=\begin{bmatrix}\dot{p}I_x-\dot{r}I_{xz}+qr\left(I_z-I_y\right)-pqI_{xz}\\\dot{q}I_y+pr\left(I_x-I_z\right)+\left(p^2-r^2\right)I_{xz}\\\dot{r}I_z-\dot{p}I_{xz}+pq\left(I_y-I_x\right)+qrI_{xz}\end{bmatrix}
\end{gathered}
\end{equation}

\subsubsection{Kinematic Equations}
The relationship between the attitude angular rates and the angular velocity components in the body frame:

\begin{equation}
\begin{bmatrix}p\\q\\r\end{bmatrix} =
\begin{bmatrix}
\dot\phi - \dot\psi \sin\theta \\
\dot\theta \cos\phi + \dot\psi \cos\theta \sin\phi \\
- \dot\theta \sin\phi + \dot\psi \cos\theta \cos\phi
\end{bmatrix} 
\end{equation}

The weight of the airship is much greater than the maximum thrust that the thruster can provide, $G \gg T_{\text{max}}$, so it can be approximated that the airship is sufficiently stable in pitch and roll, and thus always maintains $\theta = \phi = 0$. The kinematic equations of the airship can be simplified as follows: 

\begin{equation}
\begin{bmatrix}\dot{x_g}\\\dot{y_g}\\\dot{-h}\end{bmatrix}=\boldsymbol{\underset{bg}{L}}^T\begin{bmatrix}u\\v\\w\end{bmatrix}
\end{equation}

\subsection{Simplified Airship Dynamics}
The weight of the airship is much greater than the maximum thrust that the thruster can provide, $G \gg T_{\max}$. Therefore, the airship can be approximated as sufficiently stable in pitch and roll and can always maintain $\theta = \phi = 0$. The kinematic equations of the airship are simplified as follows: 

\begin{equation}
\boldsymbol{L}_{bg} =
\begin{bmatrix}
\cos\psi & \sin\psi & 0\\
-\sin\psi & \cos\psi & 0\\
0 & 0 & 1
\end{bmatrix} 
\end{equation}

\begin{equation}
\begin{bmatrix}p\\q\\r\end{bmatrix} =
\begin{bmatrix}0\\0\\\dot{\psi}\end{bmatrix} 
\end{equation}

\begin{equation}
\begin{bmatrix}\dot{u}\\\dot{v}\\\dot{w}\end{bmatrix} =
\begin{bmatrix}vr\\-ur\\0\end{bmatrix} +
\dfrac{1}{m} \Bigg(
\boldsymbol{L}_{ba} \begin{bmatrix}-D\\ \mathbf{0} \\ -L\end{bmatrix} +
\begin{bmatrix}T \cos\delta_y \cos\delta_p\\ T \sin\delta_y \cos\delta_p\\ -T \sin\delta_p\end{bmatrix}
\Bigg) 
\end{equation}

\begin{equation}
M \sin\beta + s_x T \sin\delta_y \cos\delta_p - C_2 r = \dot{r} I_z 
\end{equation}

\begin{equation}
\begin{bmatrix}\dot{x_g}\\\dot{y_g}\\-\dot{h}\end{bmatrix} =
\begin{bmatrix}u \cos\psi - v \sin\psi\\ u \sin\psi + v \cos\psi\\ w\end{bmatrix} 
\end{equation}

\section{Controller Design}
\subsection{Open-Loop Control}
The PUB has three controllable input variables: the output thrust of the plasma thruster $T$, and the Euler angles of the plasma thruster provided by the gimbal, $\delta_y$ and $\delta_p$.  
In open-loop mode, control signals sent remotely from a transmitter or an upper-level computer are received by the onboard ESP32-S3 microcontroller and processed according to the corresponding interface protocol. The flight control program finally normalizes the control signals into the above three input variables.

To adjust the plasma thruster output thrust $T$, the microcontroller writes a PWM signal, whose pulse width corresponds to the throttle control signal, to the corresponding GPIO port. The GPIO port is connected to an adjustable voltage regulator module and a boost module, thereby applying an appropriate high voltage to the thruster to obtain the desired thrust output.

To adjust the plasma thruster Euler angles $\delta_y$ and $\delta_p$, the microcontroller writes a PWM signal, whose pulse width corresponds to the heading control signal, to the corresponding GPIO port. This port is connected to the servos controlling the two degrees of freedom of the gimbal, thus controlling the angular displacement output of the servos to achieve the desired thruster orientation Euler angles.

\subsection{Inner-Loop Stabilization about a Trim}
To achieve attitude stability during the PUB’s motion process, it is necessary to design an inner-loop attitude controller.  
According to the system dynamics model, we have~\eqref{eq:29}:  

\begin{equation}
\begin{bmatrix}\dot{u}\\\dot{v}\\\dot{w}\end{bmatrix}
=
\begin{bmatrix}vr\\-ur\\0\end{bmatrix}
+\frac{1}{m}\Bigg(
\boldsymbol{L}_{\boldsymbol{b}a}
\begin{bmatrix}-D\\\boldsymbol{0}\\-L\end{bmatrix}
+\begin{bmatrix}T\cos\delta_y\cos\delta_p\\T\sin\delta_y\cos\delta_p\\-T\sin\delta_p\end{bmatrix}
\Bigg)
\label{eq:29}
\end{equation}

\begin{equation}
M\sin\beta + s_xT\sin\delta_y\cos\delta_p - C_2r = \dot{r}I_z
\end{equation}

It can be observed that the model exhibits significant nonlinearity, so linearization is attempted.  
Let the reference flight state be low-speed level flight, and define the output deviations as  
$\vartriangle u = u - u_0,\ \vartriangle v = v - v_0,\ \vartriangle w = w - w_0,\ \vartriangle r = r - r_0$,  
and the input deviations as  
$\vartriangle T = T - T_0,\ \Delta\delta_y = \delta y - \delta y_0,\ \Delta\delta_p = \delta p - \delta p_0$.  

From the reference flight state, the initial values are given by $\delta y_0 = 90^\circ,\ \delta p_0 = 0^\circ$.  
Considering that the disturbances during the motion process are generally small, all deviations can be regarded as small quantities, and the products of deviations are second-order small quantities that can be neglected.  

By introducing the small-angle assumption and applying a Taylor expansion, the linearized state-space~\eqref{eq:31} can be further obtained:  
\begin{equation}
\frac{d}{dt}
\begin{bmatrix}
\Delta u\\
\Delta v\\
\Delta w\\
\Delta r
\end{bmatrix}
=
\begin{bmatrix}
-\frac{1}{m}\rho VC_D & 0 & 0 & 0 \\
0 & \frac{1}{2m}\rho VC_{L\alpha} & 0 & -V \\
0 & 0 & \frac{1}{2m}\rho VC_{L\alpha} & 0 \\
0 & \frac{1}{2m}\rho Vc_0C_{m\alpha} & 0 & -\frac{C_2}{l_z}
\end{bmatrix}
\begin{bmatrix}
\Delta u\\
\Delta v\\
\Delta w\\
\Delta r
\end{bmatrix}
+
\begin{bmatrix}
\frac{1}{m} & 0 & 0 \\
0 & \frac{T}{m} & 0 \\
0 & 0 & -\frac{T}{m} \\
0 & \frac{s_xT}{l_z} & 0
\end{bmatrix}
\begin{bmatrix}
\Delta T\\
\Delta\delta_y\\
\Delta\delta_p
\end{bmatrix}
\label{eq:31}
\end{equation}

Starting from~\eqref{eq:31}, first consider

\begin{equation}
\Delta \dot{u} = -\frac{1}{m}\rho V C_D \Delta u + \frac{1}{m} \Delta T
\end{equation}

in order to stabilize $\Delta u$, design the state feedback

\begin{equation}
\Delta T = -k_u \Delta u + \Delta T_1
\end{equation}

where $k_u > 0$ and $\Delta T_1$ is the external input to the system.  
Similarly, for

\begin{equation}
\Delta \dot{w} = \frac{1}{2m}\rho V C_{L\alpha} \Delta w - \frac{T}{m} \Delta \delta_p
\end{equation}

design the state feedback

\begin{equation}
\Delta \delta_p = k_w \Delta w + \Delta \delta_{p_1}
\end{equation}

According to the stability condition 

\begin{equation}
k_w > \frac{\rho V C_{L\alpha}}{2T}
\end{equation}

The restriction on the proportional gain $k_w$ is obtained as:

\begin{equation}
k_w > \frac{\rho V C_{L\alpha}}{2T}
\end{equation}

The remaining two channels are described by~\eqref{eq:38}:

\begin{equation}
\begin{bmatrix}
\dot{\Delta v} \\
\dot{\Delta r}
\end{bmatrix}
=
\begin{bmatrix}
\frac{\rho V C_{L\alpha}}{2m} & -V \\
\frac{\rho V c_0 C_{M\alpha}}{2m} & -\frac{C_2}{I_z}
\end{bmatrix}
\begin{bmatrix}
\Delta v \\
\Delta r
\end{bmatrix}
+
\begin{bmatrix}
\frac{T}{m} \\
\frac{S_x T}{I_z}
\end{bmatrix}
\Delta \delta_y
\label{eq:38}
\end{equation}

Based on the form of the state equation, state feedback control is still adopted.  
When the system has zero external input, let

\begin{equation}
\Delta \delta_y = -k_1 \Delta v - k_2 \Delta r
\end{equation}

substituting into~\eqref{eq:38} yields:
\begin{equation}
\begin{bmatrix}
\dot{\Delta v} \\
\dot{\Delta r}
\end{bmatrix}
=
\begin{bmatrix}
\frac{\rho V C_{L\alpha}}{2m} - \frac{T}{m} k_1 & -V - \frac{T}{m} k_2 \\
\frac{\rho V c_0 C_{M\alpha}}{2m} - \frac{S_x}{I_z} k_1 & -\frac{C_2}{I_z} - \frac{S_x}{I_z} k_2
\end{bmatrix}
\begin{bmatrix}
\Delta v \\
\Delta r
\end{bmatrix}
\label{eq:40}
\end{equation}

Let

\begin{equation}
A =
\begin{bmatrix}
\frac{\rho V C_{L\alpha}}{2m} - \frac{T}{m} k_1 & -V - \frac{T}{m} k_2 \\
\frac{\rho V c_0 C_{M\alpha}}{2m} - \frac{S_x}{I_z} k_1 & -\frac{C_2}{I_z} - \frac{S_x}{I_z} k_2
\end{bmatrix}
\end{equation}

According to Lyapunov’s second method, the proportional gains $k_1$ and $k_2$ can be determined such that the equilibrium point

\begin{equation}
x = \begin{bmatrix} \Delta v \\ \Delta r \end{bmatrix} = \begin{bmatrix} 0 \\ 0 \end{bmatrix}
\end{equation}

of the system defined by~\eqref{eq:40} is asymptotically stable.  
Construct a symmetric matrix

\begin{equation}
M =
\begin{bmatrix}
m_1 & m_2 \\
m_2 & m_4
\end{bmatrix}
\end{equation}

such that

\begin{equation}
A^T M + M A = -I
\label{eq:44}
\end{equation}

where $I$ is the $2 \times 2$ identity matrix, which is clearly positive definite.  
By solving~\eqref{eq:44}, the matrix $M$ can be obtained.  
If $k_1$ and $k_2$ are designed such that $M$ is positive definite, a positive definite energy function

\begin{equation}
V(x) = x^T M x
\end{equation}

can be constructed, which satisfies

\begin{equation}
\dot{V}(x) = -x^T I x
\end{equation}

being negative definite.  
Therefore, the system is asymptotically stable.  
In summary, to ensure attitude stability, the control inputs are designed as:

\begin{equation}
\Delta T = -k_u \Delta u
\end{equation}

\begin{equation}
\Delta \delta_p = k_w \Delta w
\end{equation}

\begin{equation}
\Delta \delta_y = -k_1 \Delta v - k_2 \Delta r
\end{equation}

where $k_u > 0$, $k_w > \frac{\rho V C_{L\alpha}}{2T}$, and $k_1$, $k_2$ satisfy the constraints imposed by Lyapunov’s second method.

\begin{figure}[htbp]
  \centering
  \includegraphics[width=0.75\textwidth]{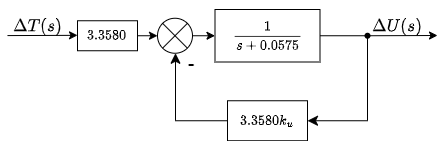}
  \caption{Transfer function model of controller.}
  \label{fig:fig14}
\end{figure}

Below, the parameter tuning is illustrated using $\Delta u$ as an example.  
According to the finite element calculations of the PUB aerodynamic characteristics, we have $C_D = 0.0071$. As previously mentioned, the mass is $m = 0.2978$~kg, and the air density is taken as $\mathfrak{p} = 1.205$~kg/m$^3$ (all parameters in SI units).  
The flight speed is set to 1~m/s.  
By applying the Laplace transform to the equation with the introduced state feedback, the transfer function model is obtained as shown in Fig.~\ref{fig:fig14}.

\begin{figure}[htbp]
  \centering
  \includegraphics[width=0.75\textwidth]{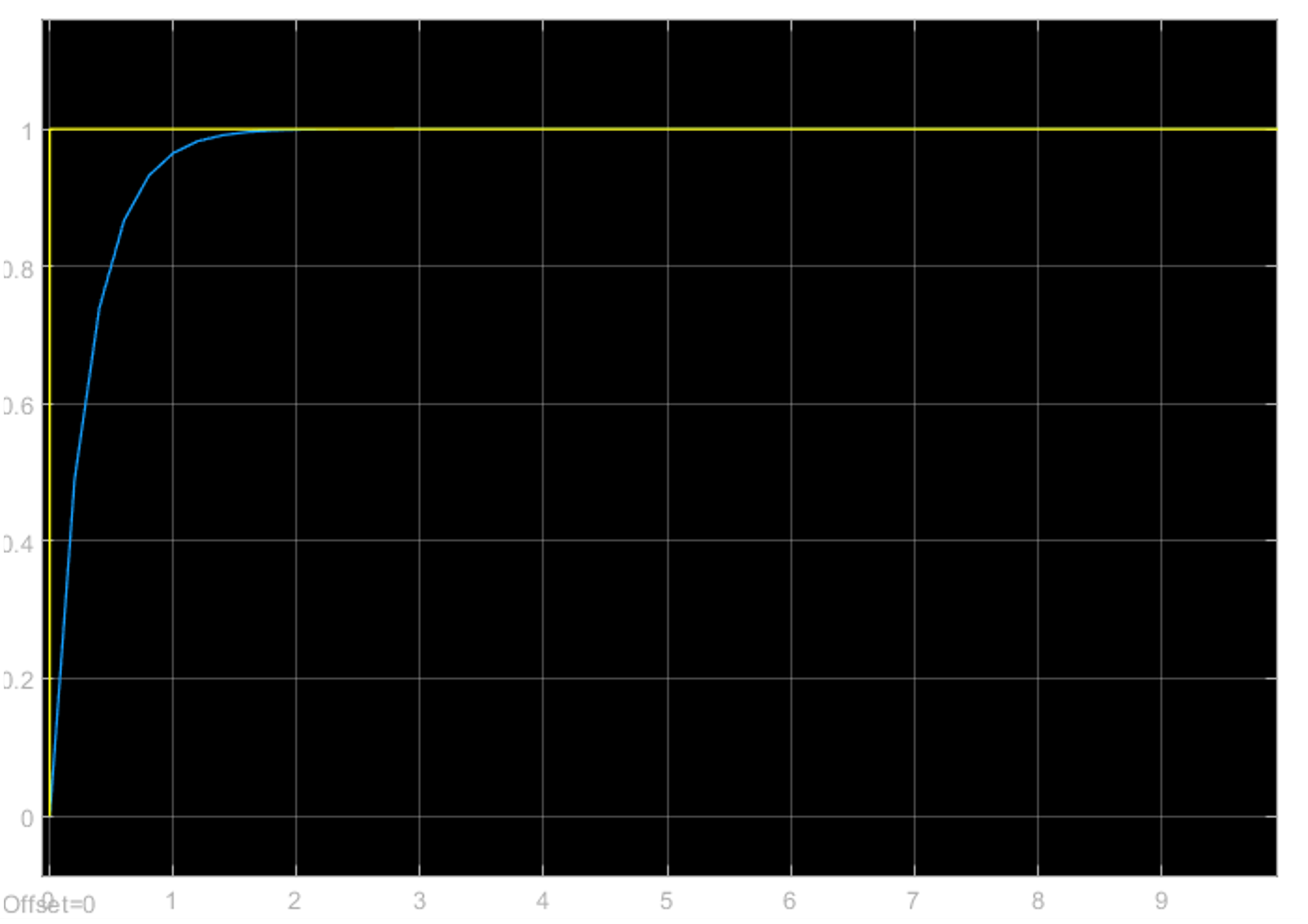}
  \caption{Transfer function model of controller.}
  \label{fig:fig15}
\end{figure}

The closed-loop transfer function of the system is
\begin{equation}
\Phi(s) = \frac{3.3580}{s + 0.0575 + 3.3580 k_u} \label{eq:transfer}
\end{equation}
To eliminate the steady-state error of the system in response to a step input, the gain is designed as $k_u = 0.9828$.  
A simulation is built in MATLAB/Simulink, and the system step response obtained is shown in Fig.~\ref{fig:fig15}.  
By analyzing the response curve, it can be seen that the controller has strong tracking performance and good steady-state characteristics.

\subsection{Sliding-Mode Control (SMC) for Attitude Tracking}

Under the above assumptions, we linearized the motion states of the PUB by setting the reference motion to low-speed level flight.  
Based on the linearized model, previous studies have proposed control methods such as PID controllers~\cite{alsayed2017}, linear matrix inequality (LMI) methods, and boundary layer sliding mode control methods.  
However, in actual flight experiments, the motion states of the PUB are affected by nonlinear factors such as airflow disturbances and self-induced vibrations.  
Therefore, based on the attitude tracking sliding mode control method proposed in~\cite{yang2012}, we further adopt an optimized control technique for the PUB.  

According to~\cite{xixiang2019}, we substitute the lateral-direction linearized dynamics equations into the present system:

\begin{equation}
\mathbf{M}_{s} \dot{\mathbf{X}} = \mathbf{A}_{s} \mathbf{X} + \mathbf{B}_{s} \mathbf{U}
\label{eq:51}
\end{equation}

\begin{equation}
\mathbf{M}_{s} =
\begin{bmatrix}
m + m_{11} & 0 & -m y_{G} \\
m + m_{22} & 0 & m \cdot x_{G} \\
-m \cdot y_{G} & m \cdot x_{G} & I_{z} + m_{66}
\end{bmatrix}
\label{eq:Ms}
\end{equation}

\begin{equation}
\mathbf{A}_{s} =
\begin{bmatrix}
X_{a}^{u} & X_{a}^{r} & X_{a}^{r} + (m_{11} - m_{22}) v_{w} \\
Y_{a}^{u} & Y_{a}^{v} & Y_{a}^{r} + (m_{11} - m_{22}) u_{w} \\
N_{a}^{u} & N_{a}^{v} & N_{a}^{r}
\end{bmatrix}
\label{eq:As}
\end{equation}

\begin{equation}
\mathbf{B}_{s} =
\begin{bmatrix}
1 & 0 & 0 \\
0 & 1 & 0 \\
0 & 0 & 1
\end{bmatrix}, \quad
\mathbf{U} =
\begin{bmatrix}
T \\
0 \\
N_{c, r}
\end{bmatrix}, \quad
\mathbf{X} =
\begin{bmatrix}
u \\
v \\
r
\end{bmatrix}
\label{eq:BsUsXs}
\end{equation}

Here, $m_{11}$, $m_{22}$, and $m_{66}$ are the components of the added mass matrix $\mathbf{M}_a$.  
$X_{a}^{u}, X_{a}^{r}, X_{a}^{r} + (m_{11} - m_{22}) v_{w}$ are axial aerodynamic derivatives;  
$Y_{a}^{u}, Y_{a}^{v}, Y_{a}^{r} + (m_{11} - m_{22}) u_{w}$ are lateral aerodynamic derivatives;  
$N_{a}^{u}, N_{a}^{v}, N_{a}^{r}$ are yawing moment derivatives;  
$v_{w}, u_{w}$ are the wind speed components, and $T$ is the thruster force.

Sliding mode control is a robust control method. The principle is as follows: first, define a sliding surface, then drive the system trajectory to approach the sliding surface. Once near the sliding surface, the control ensures that the trajectory slides along the surface and eventually reaches it, and then moves along the surface until the system attains the desired state~\cite{xixiang2019}.

The design steps for sliding mode control are: first, design the switching function $s(x)$, then design the sliding mode control law by specifying control inputs $u^{+}(x)$ and $u^{-}(x)$ so that the system satisfies the reaching condition.

(1) First, design the sliding surface equation:
\begin{equation}
s(x) = c_1 \mathbf{e} + c_2 \dot{\mathbf{e}}, \quad (c_1, c_2 \ \text{are constants})
\label{eq:slidingsurf}
\end{equation}

For low-altitude flight of the PUB, define the tracking error as the difference between the current actual trajectory and the desired trajectory:

\begin{equation}
\mathbf{e} = \boldsymbol{\eta} - \boldsymbol{\eta}_d =
\begin{bmatrix}
x_g - x_e \\
y_g - y_e \\
\psi - \psi_e
\end{bmatrix}
\label{eq:56}
\end{equation}

where $x_g$, $y_g$, $x_e$, $y_e$ are coordinates in the ground inertial frame, and $\psi$, $\psi_e$ are the yaw angles.  

To eliminate chattering in the sliding model control, an exponential reaching law is chosen:

\begin{equation}
\dot{s} = -\varepsilon \operatorname{sgn}(s) - k s, \quad (k > 0)
\label{eq:57}
\end{equation}

Combining~\eqref{eq:51} and~\eqref{eq:56} yields the following:

\begin{equation}
\varepsilon \operatorname{sgn}(s) + k s + c_1 \dot{\mathbf{e}} + c_2 \ddot{\mathbf{e}} = 0
\label{eq:combine1}
\end{equation}

Substituting the second derivative of the error gives the following:

\begin{equation}
\varepsilon \operatorname{sgn}(s) + k \left( c_1 \mathbf{e} + c_2 \dot{\mathbf{e}} \right)
+ c_1 \dot{\mathbf{e}} + c_2 \ddot{\mathbf{e}} = 0
\label{eq:59}
\end{equation}

The transformation matrix from the ground inertial frame to the body frame is $\mathbf{C}_b^g$, which gives the relationship between the state variables and the actual trajectory $\boldsymbol{\eta}$.

\begin{equation}
\dot{\mathbf{X}} = \dot{\mathbf{C}}_b^g \dot{\boldsymbol{\eta}}
\label{eq:60}
\end{equation}

Differentiating~\eqref{eq:60} gives the following:

\begin{equation}
\dot{\mathbf{X}} = \dot{\mathbf{C}}_b^g \ddot{\boldsymbol{\eta}} + \ddot{\mathbf{C}}_b^g \dot{\boldsymbol{\eta}}
\label{eq:61}
\end{equation}

Combining~\eqref{eq:51},~\eqref{eq:60} and~\eqref{eq:61}, we obtain the following.

\begin{equation}
\ddot{\boldsymbol{\eta}} =
(\mathbf{C}_b^g)^{-1}
\left( \mathbf{M}_s^{-1} \mathbf{A}_s \mathbf{C}_b^g \dot{\boldsymbol{\eta}}
+ \mathbf{M}_s^{-1} \mathbf{B}_s \mathbf{U}
- \dot{\mathbf{C}}_b^g \dot{\boldsymbol{\eta}} \right)
\label{eq:ddeta}
\end{equation}

Substituting into~\eqref{eq:59}, the control input is obtained as:

\begin{equation}
\mathbf{U} =
\mathbf{B}_s^{-1} \mathbf{M}_s \dot{\mathbf{C}}_b^g
\left\{ \dot{\boldsymbol{\eta}}
- \mathbf{M}_s^{-1} \mathbf{A}_s \dot{\boldsymbol{\eta}}
- \frac{1}{c_2} \left[ \varepsilon \operatorname{sgn}(s)
+ k(c_1 \mathbf{e} + c_2 \dot{\mathbf{e}}) + c_1 \dot{\mathbf{e}} \right] \right\}
\label{eq:control}
\end{equation}

(2) Stability Analysis:

According to the system stability criterion—for an equilibrium point $s$, if there exists a continuous function $V$ such that:
\begin{equation}
\lim_{|s| \to \infty} V = \infty
\end{equation}
and when $s \neq 0$,
\begin{equation}
\dot{V} < 0
\end{equation}
then the system is stable at $s = 0$~\cite{shevitz1994}.  

Construct the Lyapunov function:
\begin{equation}
V = \frac{1}{2} s^2
\label{eq:66}
\end{equation}
\begin{equation}
\dot{V} = s \dot{s}
\label{eq:lyapdot}
\end{equation}

Using the reaching law designed in~\eqref{eq:57}, it is clear that~\eqref{eq:66} satisfies the first condition, and:
\begin{equation}
\dot{V} = - \varepsilon |s| - k s^2 < 0, \quad (k > 0)
\label{eq:stability}
\end{equation}
thus satisfying the second condition.  

In conclusion, the closed-loop sliding mode controller based on flight attitude for this system exhibits good stability and robustness.

\section{Prototype Experiments}
Regarding flight parameters, the feasibility experiment for the PUB design was conducted in a near-ground space. Markers were attached to the front and rear ends of the airship, and low-altitude motion videos were recorded from a fixed side view. The images can be further processed to study the motion characteristics. In actual measurements, due to insufficient stiffness of the two-degree-of-freedom gimbal base connection, the thruster yaw angle may exhibit small oscillations. Over time, these accumulate and deviate from the initial yaw angle. In the future, the gimbal connection structure will be optimized to reduce the PUB yaw angle error.  

\begin{figure}[htbp]
  \centering
  \includegraphics[width=0.75\textwidth]{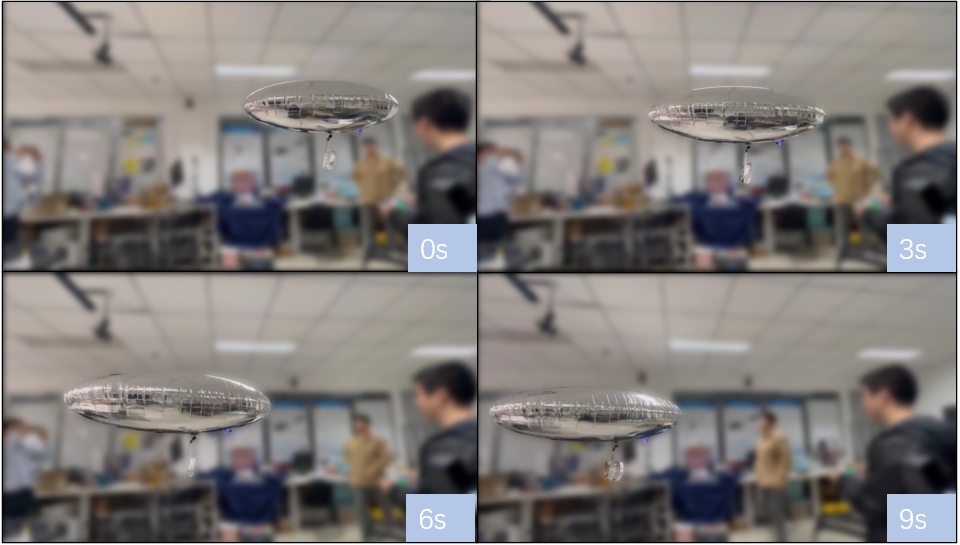}
  \caption{Side-view motion capture frames of PUB.}
  \label{fig:fig16}
\end{figure}

Fig.~\ref{fig:fig16} shows the speed measurement experiment of the PUB. By capturing four sets of flight position images at 3-second intervals, the maximum flight speed of the PUB was calculated to be 0.32~m/s. The initial flight altitude was 1.8~m, and after 12 seconds, the altitude decreased to 1.6~m, demonstrating good low-altitude flight stability.  

From side-view motion capture frames at \SI{3}{s} intervals, PUB's maximum speed is about \SI{0.32}{m/s}. The initial altitude is \SI{1.8}{m} and decreases to \SI{1.6}{m} after \SI{12}{s}, indicating stable low-altitude flight. Front-view frames confirm that the gimbal responds rapidly to yaw commands. During flight, measured sound levels ranged from \SI{55}{dB} to \SI{65}{dB}, demonstrating ultra-quiet operation.

\begin{figure}[htbp]
  \centering
  \includegraphics[width=0.75\textwidth]{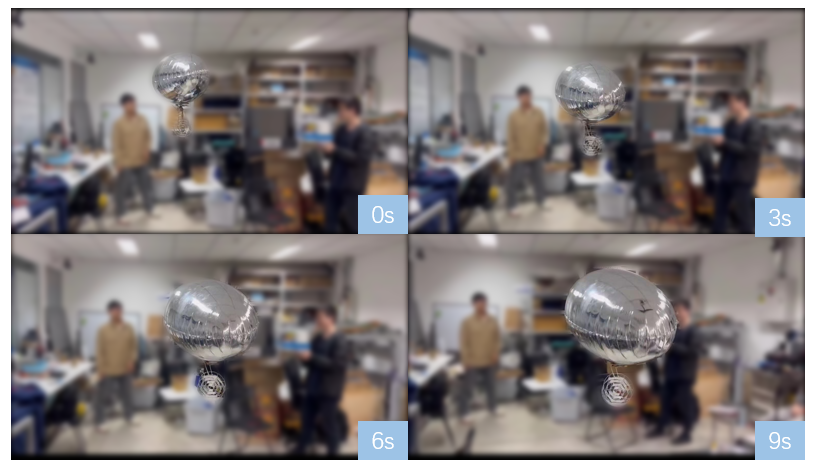}
  \caption{Front-view motion capture frames of PUB.}
  \label{fig:fig17}
\end{figure}

Fig.~\ref{fig:fig17} shows the motion capture images for the yaw verification experiment of the PUB. In the experiment, a fixed viewpoint facing along the body axis $Ox_{a}$ was used for recording. From the four motion capture images taken at 3-second intervals in  Fig.~\ref{fig:fig14}, it can be seen that when the rudder on the remote controller was pushed to the right at 0~s, the servo on the two-degree-of-freedom gimbal was able to quickly adjust the yaw angle. The above experiment demonstrates that the remote controller can effectively control both the yaw and pitch angles of the PUB.  

In addition, to verify the ultra-quiet performance of the PUB, a sound level meter was used to measure and record the noise during the aforementioned flight. According to the readings, the sound level ranged between 55~dB and 65~dB, demonstrating good quietness.  

\section{Conclusion}
We presented PUB, a plasma-propelled, ultra-quiet, low-altitude flying robot. Compared with conventional electric propulsors, PUB uses ionic wind as the core thrust mechanism, providing low mass, low noise, and controllable thrust without mechanical rotors. Modular plasma units can be combined for different missions to achieve highly maneuverable aerostat control. A two-DOF gimbal enables precise direction and altitude control. Sliding-mode control enhances robustness to disturbances. Experiments validated basic low-altitude flight capability, with speeds up to \SI{0.32}{m/s} near \SI{1.7}{m} altitude and noise levels of \SIrange{55}{65}{dB}.
Future work will add a second thruster placed symmetrically to decouple yaw and pitch, improving controllability and stability; upgrade structural materials to reduce vibration; and explore near-space applications, where ionic-wind-propelled airships may have advantages in maneuverability.


\begin{thebibliography}{99}
\bibitem{govreport2024}
People's Daily. ``Low-altitude economy written into the Government Work Report for the first time,'' China Government Network, 2024-04-02. (in Chinese)

\bibitem{whitepaper2019}
Sun, X., Cheng, W., Mu, Z., \emph{et al.} ``White Paper on the Development of Electric Aircraft,'' \emph{Aviation Science and Technology}, 30(11), 1--7, 2019. (in Chinese)

\bibitem{low_noise_propeller}
Huang, Z., Yao, H., Lundbladh, A., Davidson, L. ``Low-noise propeller design for quiet electric aircraft,'' \emph{AIAA Aviation 2020 Forum}, p.~2596, 2020.

\bibitem{brown_effect_analysis}
Ianconescu, R., Sohar, D., Mudrik, M. ``An analysis of the Brown--Biefeld effect,'' \emph{Journal of Electrostatics}, 69(6), 512--521, 2011.

\bibitem{xu2018nature}
Xu, H., He, Y., Strobel, K. L., \emph{et al.} ``Flight of an aeroplane with solid-state propulsion,'' \emph{Nature}, 563(7732), 532--535, 2018.

\bibitem{zhang2023cjoa}
Zhang, H., Leng, J., Liu, Z., Qi, M., Yan, X. ``Passive attitude stabilization of ionic-wind-powered micro air vehicles,'' \emph{Chinese Journal of Aeronautics}, 36(7), 412--419, 2023.

\bibitem{taccogna2004}
Taccogna, F., Longo, S., Capitelli, M. ``Plasma-surface interaction model with secondary electron emission effects,'' \emph{Physics of Plasmas}, 11(3), 1220--1228, 2004.

\bibitem{hinton1983}
Hinton, F. L. ``Collisional transport in plasma,'' \emph{Handbook of Plasma Physics}, vol.~1, 331, 1983.

\bibitem{sutherland1905}
Sutherland, W. ``A dynamical theory of diffusion for non-electrolytes,'' \emph{Philosophical Magazine}, 9(54), 781--785, 1905.

\bibitem{wilson2009}
Wilson, J., Perkins, H. D., Thompson, W. K. ``An investigation of ionic wind propulsion,'' Technical Report, 2009.

\bibitem{chang1991}
Chang, J. S., Lawless, P. A., Yamamoto, T. ``Corona discharge processes,'' \emph{IEEE Transactions on Plasma Science}, 19(6), 1152--1166, 1991.

\bibitem{ahedo2011}
Ahedo, E. ``Plasmas for space propulsion,'' \emph{Plasma Physics and Controlled Fusion}, 53(12), 124037, 2011.

\bibitem{xu_thesis_2020}
Xu, H. \emph{Experiments in Electroaerodynamic Propulsion}, Ph.D. Thesis, MIT, 2020.

\bibitem{alsayed2017}
Alsayed, A., Lanteigne, E. ``Experimental pitch control of an unmanned airship with sliding ballast,'' \emph{2017 International Conference on Unmanned Aircraft Systems (ICUAS)}, 1640--1646.

\bibitem{yang2012}
Yang, Y., Wu, J., Zheng, W. ``Terminal sliding mode control for attitude tracking of an autonomous airship,'' 2012, 32(4), 29--36. (in Chinese)

\bibitem{xixiang2019}
Yang, X., Zhang, J. ``Sliding-mode control for stratospheric airship trajectory tracking in wind fields,'' \emph{Journal of National University of Defense Technology}, 41(1), 2019. (in Chinese)

\bibitem{shevitz1994}
Shevitz, D., Paden, B. ``Lyapunov stability theory of nonsmooth systems,'' \emph{IEEE Transactions on Automatic Control}, 39(9), 1910--1914, 1994.

\bibitem{majiaxing2021}
Ma, J., Quan, R., Teng, H., \emph{et al.} ``Trajectory analysis of stratospheric airships based on ionic-wind electric propulsion,'' \emph{Acta Astronautica Sinica}, 42(5), 642--649, 2021. (in Chinese)
\end{thebibliography}
\end{document}